\documentclass[preprint,12pt]{elsarticle}

\usepackage{algorithm}
\usepackage{algpseudocode}
\usepackage{graphicx}
\usepackage{xcolor}
\usepackage{textcomp}
\usepackage{amsmath,amssymb,graphicx,booktabs}
\usepackage{caption}
\usepackage{subcaption}
\usepackage{multirow}
\usepackage{hyperref}
\usepackage{lineno}

\modulolinenumbers[5]

\journal{ArXiv}

\begin{document}

\begin{frontmatter}



\title{Metabolomic Biomarker Discovery for ADHD Diagnosis Using Interpretable Machine Learning}


\author{Nabil Belacel} 

\affiliation{organization={Digital Technology Research Center, National Research Council Canada},
            city={Ottawa},
            state={Ontario},
            country={Canada}}
\author{Mohamed Rachid Boulassel}

\affiliation{organization={Biomedical Science and Hematology Departments, College of Medicine and Health Sciences, Sultan Qaboos University},
            city={Muscat},
            country={Sultanate of Oman}}
\begin{abstract}
Attention-Deficit/Hyperactivity Disorder (ADHD) is a prevalent neurodevelopmental disorder with limited objective diagnostic tools, highlighting the urgent need for objective, biology-based diagnostic frameworks in precision psychiatry. We integrate urinary metabolomics with an interpretable machine-learning framework to identify biochemical signatures associated with ADHD. Targeted metabolomic profiles from 52 ADHD and 46 control participants were analyzed using a Closest Resemblance (CR) classifier with embedded feature selection. The CR model outperformed Random Forest and K-Nearest Neighbor classifiers, achieving an AUC > 0.97 based on a reduced panel of 14 metabolites. These metabolites—including dopamine 4-sulfate, N-acetylaspartylglutamic acid, and citrulline—map to dopaminergic neurotransmission and amino acid metabolism pathways, offering mechanistic insight into ADHD pathophysiology. The CR classifier’s transparent decision boundaries and low computational cost support integration into targeted metabolomic assays and future point-of-care diagnostic platforms. Overall, this work demonstrates a translational framework combining metabolomics and interpretable machine learning to advance objective, biologically informed diagnostic strategies for ADHD.
\end{abstract}










\end{frontmatter}



\section{Introduction}
Attention-Deficit/Hyperactivity Disorder (ADHD) is a complex neurodevelopmental condition characterized by persistent patterns of inattention, hyperactivity, and impulsivity that interfere with academic, social, and behavioral functioning \cite{shafiullah2025current}. Despite decades of clinical research, ADHD diagnosis continues to rely largely on subjective behavioral assessments, parent/teacher questionnaires, and clinician judgment methods that remain vulnerable to observer bias, cultural differences, and inter-rater variability. This diagnostic subjectivity has intensified the need for objective, biologically informed tools capable of improving accuracy and reproducibility in clinical practice \cite{zhao2025artificial, salmeron2025}.

Biomarker discovery has therefore become a central priority in ADHD research. Among emerging approaches, metabolomic the comprehensive profiling of small molecules reflecting downstream products of gene expression and neurotransmitter metabolism—offers a direct window into biochemical dysregulation associated with ADHD \cite{predescu2024metabolomic}. Urine-based metabolomic analysis is particularly advantageous: it is non-invasive, scalable, child-friendly, and sensitive to metabolic alterations linked to neurotransmission, amino acid turnover, and oxidative stress \cite{hung2025identifying, salmeron2025}. However, metabolomics datasets are typically high-dimensional, noisy, and characterized by strong feature correlations, making the identification of reliable biomarkers a challenging task \cite{tian2022urinary}.

Machine learning (ML) has emerged as an essential tool for addressing these challenges. By modeling nonlinear relationships and detecting subtle multivariate patterns, ML algorithms enable the extraction of informative biomarkers from complex metabolic profiles \cite{chen2024metabolomic, Belacel2019}. Yet traditional classifiers such as Support Vector Machines (SVM), Random Forests (RF), logistic regression, and k-Nearest Neighbors (kNN) often struggle with issues of interpretability, redundancy among features, and computational demands. These limitations hinder clinical translation, particularly when the goal is to derive small, robust metabolite panels suitable for diagnostic assays \cite{Gong2023}.

Machine-learning–based biomarker discovery often relies on filter, wrapper, or embedded feature selection methods that prioritize numerical optimization over biological interpretability \cite{sid2025efficient,cuperlovic2010multi}. These approaches can yield metabolite subsets that are statistically discriminative yet mechanistically ambiguous, and post hoc explanation tools such as SHAP or LIME provide only indirect and sometimes unstable interpretations \cite{guldogan2025interpretable}. Metabolomic data further complicate feature selection due to high dimensionality, correlated pathways, and measurement variability.

To address these limitations, we introduce a Closest Resemblance (CR) classifier that integrates interval feature learning with an outranking-based similarity measure. By representing each metabolite as an interval rather than a point estimate, the CR model inherently accounts for biological variability and assay uncertainty, improving robustness to noise and non-Gaussian distributions \cite{Belacel2000,belacel2004k,Belacel2019}. The resulting decision process is intrinsically interpretable, as class assignments arise directly from metabolite-level resemblance relations rather than post hoc surrogate explanations \cite{ng2025explainable}. This framework offers a principled and transparent alternative to conventional ML methods for metabolomic biomarker discovery.
The objectives of this study are: (1) to evaluate the performance of the CR classifier on a publicly available urinary metabolomic ADHD dataset \cite{tian2022urinary}; (2) to identify metabolite subsets most relevant for distinguishing ADHD from controls; and (3) to benchmark CR performance against standard machine learning models. By integrating biochemical relevance with computational rigor, this work contributes to the development of interpretable, non-invasive, and scalable tools for objective ADHD diagnosis and monitoring.
\section{Related work}
\label{Related work}
\subsection{Traditional Diagnostic Practices for ADHD}
Attention-Deficit/Hyperactivity Disorder (ADHD) is typically diagnosed through clinical evaluation based on standardized behavioral criteria such as the DSM-5 and ICD-11 guidelines \citep{doernberg2016neurodevelopmental, world2024clinical}. These assessments rely heavily on clinician judgment, patient interviews, and reports from parents or teachers. Although widely used, these methods face challenges including subjectivity, inter-rater variability, cultural bias, and limited ability to differentiate ADHD from comorbid conditions (e.g., anxiety, learning disorders) \citep{furman2005attention,zhao2025artificial}.

\subsection{Need for Objective and Quantifiable Biomarkers}
The limitations of interview-based diagnosis highlight the need for objective, biologically grounded diagnostic tools. Objective biomarkers could reduce diagnostic variability, enable earlier identification, and offer a more mechanistic understanding of ADHD’s neurobiological basis \citep{hurjui2025biomarkers,zhao2025artificial}. Furthermore, biomarkers could support precision medicine by enabling individualized prediction of ADHD subtypes or treatment response.

\subsection{Challenges in Neuroimaging and Neurophysiological Biomarkers}
A large body of research has sought neurobiological markers through neuroimaging modalities such as structural MRI, functional MRI (fMRI), and diffusion tensor imaging (DTI). While these methods have revealed alterations in prefrontal, striatal, and cerebellar circuits \citep{parlatini2024state,loo2024translating}, their clinical utility remains limited. High cost, restricted accessibility, long acquisition times, motion artifacts (particularly in children), and lack of reproducibility limit their feasibility for population-level screening \citep{loo2024translating}.  
Similarly, electrophysiological measures such as electroencephalography (EEG) and event-related potentials (ERP) have revealed alterations in attentional allocation and inhibitory control networks in individuals with ADHD \citep{Barry2003,Johnstone2013}. Despite these findings, EEG-based biomarkers generally exhibit modest effect sizes and substantial inter-individual variability, which limit their diagnostic precision. In addition, EEG analysis requires extensive preprocessing including artifact correction, filtering, and component decomposition, which introduces complexity and reduces reproducibility across studies. These challenges have contributed to the limited clinical translation of EEG biomarkers for ADHD \citep{Lenartowicz2018, Loo2016}.

\subsection{Emergence of Metabolomics as a Promising Direction}
Recent work has shifted toward metabolic profiling as a more accessible and biologically integrative approach. Metabolomics captures downstream effects of gene expression, including neurotransmitter turnover and energy metabolism disturbances, providing a real-time biochemical readout of the dysregulation associated with ADHD \citep{KaddurahDaouk2009, hung2025identifying}. Several studies report alterations in amino acids, catecholamine metabolites, fatty acids, and oxidative stress markers in individuals with ADHD \citep{tian2022urinary, MunozZabaleta2025}.  
Urine metabolomics, in particular, has gained attention due to its non-invasive collection, low cost, and suitability for pediatric populations. Studies such as \cite{tian2022urinary} have demonstrated promising discriminatory power using urinary metabolites, with external validation yielding AUC values above 0.85 for selected biomarker panels \citep{tian2022urinary}.

\subsection{Machine Learning for Metabolite-Based Biomarker Discovery}
Machine learning approaches have begun to be applied to metabolomic datasets in ADHD research \cite{TANG2022102209}, though such applications remain relatively nascent. For instance, in a urinary NMR-metabolomics study of children with ADHD \cite{salmeron2025} demonstrated that combining behavioral measures with metabolic features (e.g., creatine, 3-indoxylsulfate) using logistic regression substantially improved diagnostic classification. In another metabolomic profiling of ADHD by Tian et al. \cite{tian2022urinary}, a panel of urinary metabolites (including FAPy-adenine, dopamine 4-sulfate, aminocaproic acid, and asparaginyl-leucine) was used to build predictive models via Random Forest and achieved high area under the curve (AUC) values. More broadly, systematic reviews of ADHD metabolomic studies highlight that altered amino acid, neurotransmitter and fatty acid metabolism signatures are common themes, and some pilot works have started using machine learning frameworks to identify metabolite panels \citep{predescu2024metabolomic, salmeron2025}.

Despite their potential, existing methods face challenges:  
\begin{itemize}
    \item Many models operate as black boxes, limiting interpretability and clinical trust.
    \item Feature selection is often unstable, sensitive to noise, and prone to overfitting when the number of metabolites exceeds sample size.
    \item High computational complexity can hinder deployment in real-time or embedded systems.
    \item Few studies integrate feature selection directly into the classifier, reducing transparency of biomarker identification.
\end{itemize}

\subsection{Gaps and Limitations in Current Metabolite-Based Biomarker Methods}
Existing metabolomics-based ADHD studies often rely on univariate statistical testing or generic feature-selection algorithms such as LASSO or ReliefF, which do not fully capture multivariate interactions or pathway-level covariance among metabolites \citep{Liu2019ADHDMetab,tian2022urinary}. As a result, reported biomarker panels frequently show limited overlap between studies, reflecting poor robustness and potential cohort-specific effects \citep{predescu2024metabolomic}. Furthermore, most proposed metabolite biomarkers have been evaluated only within a single dataset, without external validation or cross-cohort replication, a limitation widely acknowledged across psychiatric metabolomics \citep{Castro2023PsychMetabReview}. Few studies assess computational efficiency or real-time feasibility of machine-learning models, raising concerns about scalability for clinical implementation—particularly given that high-dimensional metabolomic data can yield unstable models when not optimized for reproducibility \citep{Cao2020MetabML}. These limitations underscore the need for more rigorous biomarker-development frameworks tailored to metabolomics data and validated in independent ADHD populations.

\subsection{Contribution of the Present Work}
To address these gaps, the present study introduces a Closest Resemblance (CR) classifier equipped with an embedded greedy feature selection strategy\citep{belacel_alg2025,Belacel2019}. Unlike conventional classifiers, the CR model is based on interval learning and outranking similarity principles, offering transparent decision boundaries and robustness to noise and data imbalance.  
The integrated feature selection procedure identifies minimal yet highly discriminative subsets of urinary metabolites, improving both classification accuracy and biological interpretability. By producing compact metabolite signatures, the proposed approach enhances feasibility for translation into targeted LC-MS/MS diagnostic panels and edge-AI applications.

\subsection{Motivation for New Algorithmic Approaches}
Given the clinical need for objective ADHD biomarkers, the limitations of imaging-based methods, and the growing evidence for metabolic dysregulation in ADHD, there is a strong need to develop new computational frameworks capable of identifying reproducible and biologically meaningful metabolite markers. This study contributes toward meeting that need by providing a robust, interpretable, and computationally efficient biomarker discovery method grounded in metabolic pathways relevant to ADHD neurobiology.

\section{Materials and Methods}

\subsection{Dataset description}
\label{subsec:dataset}

The dataset analyzed in this study was derived from the publicly available urinary metabolomic study of attention-deficit/hyperactivity disorder ADHD in children reported by \cite{tian2022urinary}. The data consist of quantitative metabolite intensities acquired by ultra-performance liquid chromatography coupled with quadrupole time-of-flight mass spectrometry (UPLC–QTOF–MS). After data preprocessing and normalization, a total of $60$ identified urinary metabolites were retained for analysis.  

The study cohort comprises 98 participants, including 52 children clinically diagnosed with ADHD and 46 age-matched healthy controls. All subjects were medication-free and selected according to standardized diagnostic criteria \textit{DSM-V} to ensure diagnostic homogeneity \cite{tian2022urinary}. The metabolite panel encompasses amino acids, biogenic amines, organic acids, and neurotransmitter related intermediates, reflecting a broad coverage of biochemical pathways such as catecholamine metabolism, urea cycle, and tryptophan/indole derivatives \cite{predescu2024metabolomic}.  

These normalized metabolite features were used as input variables $X \in \mathbb{R}^{98\times60}$ for the classification experiments, with diagnostic classes C representing the two categories (Control vs. ADHD). The proposed Closest Resemblance CR classifier was evaluated using this dataset to assess its predictive performance and ability to identify metabolite subsets discriminative of ADHD.

\subsection{Closest Resemblance (CR) Classifier}
\label{sec:CR}
In this subsection, we briefly describe the Closest Resemblance (CR) classifier; a more detailed formulation is presented in \cite{belacel_alg2025}. The CR classifier is a supervised learning approach that integrates feature interval learning with an outranking-based similarity measure. It is conceptually derived from the PROAFTN multi-criteria classification framework introduced by \cite{Belacel2000}. The CR and PROAFTN family of classifiers have been successfully applied to a wide range of real-world problems requiring robust and interpretable decision-making under uncertainty, such as acute leukemia diagnosis \cite{belacelCMPBAcuteL2001,BelacelAIM2001}, Alzheimer’s disease detection \cite{Belacel2019}, e-Health applications \cite{belacelTMJ2005web}, recommendation systems \cite{BelacelICAART2020}, remote sensing data analysis \cite{belacelREMOTESensingData2020k}. 

By modeling each feature as an interval rather than a single numeric value, the CR classifier explicitly accounts for natural variability and measurement uncertainty in the data. This provides a more stable and interpretable decision framework compared to traditional distance-based classifiers. The CR's classification rule is: "A sample $s$ is assigned to class $C$ if and only if it is most similar, or (resembles) the prototype of that class"\citep{belacel_alg2025,Belacel2000}.

\subsubsection{Learning Phase}
\label{sec:learning}

Let a set of $L$ labeled examples $X=\{x_1, x_2, ..., x_L\}$ be defined over $n$ features $F=\{f_1, f_2, ..., f_n\}$, where each instance $x_i$ belongs to one of $K$ classes. Each instance is represented as a feature vector $f(x_i)=\langle f_1(x_i), f_2(x_i), ..., f_n(x_i)\rangle$. In the learning phase, CR constructs a prototype $P(C^h)$ for each class $C^h$, $h=1, ..., K$, representing the typical pattern of that class. The prototype is defined as a set of feature intervals:

\begin{equation}
p(C^h) = \{ I(f_1, C^h), I(f_2, C^h), ..., I(f_n, C^h) \}
\label{featureIterval}
\end{equation}

Each interval $I(f_j, C^h) = [I^1(f_j, C^h), I^2(f_j, C^h)]$ defines the range of feature $f_j$ values observed for class $C^h$. In this study, we evaluated two simple but effective approaches to construct these intervals: one based on the mean and standard deviation (Eq.~\ref{eq.meanstd}), and one based on percentile bounds (Eq.~\ref{eq.percentile}).

\begin{equation}
I_j(f_j, C^h) = [\mu(f_j, C^h) - t\times \sigma(f_j, C^h), \mu(f_j, C^h) + t\times \sigma(f_j, C^h)]
\label{eq.meanstd}
\end{equation}

\begin{equation}
I_j(f_j, C^h) = [L_j^h(f_j, C^h), U_j^h(f_j, C^h)]
\label{eq.percentile}
\end{equation}

These interval definitions are non-parametric and do not assume any specific underlying data distribution, making them suitable for heterogeneous environmental data.

\subsubsection{Classification Phase}

For a testing instance $s$, CR first computes a performance matrix representing the deviation between $s$ and the prototype of each class $C^h$ for every feature $f_j$. The absolute distance is computed as:

\begin{equation}
d_j^h(s, p(C^h)) = \max\{0, I^1_{j,h} - f_j(s), f_j(s) - I^2_{j,h}\}
\label{eq.absdistance}
\end{equation}

A value of $0$ indicates that $f_j(s)$ lies within the prototype interval, while larger values represent greater deviation. These distances are then aggregated into an outranking relation that compares the resemblance of $s$ across all class prototypes. 

The partial outranking index for each feature $f_j$ is defined as:

\begin{equation}
R_j^s(p(C^h), p(C^l)) =
\begin{cases}
1, & \text{if } d_j^h(s, p(C^h)) \leq d_j^l(s, p(C^l)) \\
0, & \text{otherwise}
\end{cases}
\label{eq.partialoutranking}
\end{equation}

Considering the importance weight $w_j$ of each feature ($\sum_{j=1}^n w_j = 1$), the global outranking index is given by:

\begin{equation}
R^s(p(C^h), p(C^l)) = \sum_{j=1}^{n} w_j R_j^s(p(C^h), p(C^l))
\label{eq.globalOutranking}
\end{equation}

The positive and negative outranking flows are then computed following the PROMETHEE decision framework ~\cite{brans2005promethee}:

\begin{equation}
\phi^+(p(C^h)) = \sum_{l=1, l \neq h}^{K} R^s(p(C^h), p(C^l))
\label{eq.positiveflow}
\end{equation}
\begin{equation}
\phi^-(p(C^h)) = \sum_{l=1, l \neq h}^{K} R^s(p(C^l), p(C^h))
\label{eq.negativeflow}
\end{equation}

The net outranking flow,
\begin{equation}
\phi(p(C^h)) = \phi^+(p(C^h)) - \phi^-(p(C^h)),
\label{eq.netflow}
\end{equation}

represents the overall resemblance score of the prototype $p(C^h)$ to the data point $s$. Finally, CR assigns $s$ to the class with the highest net flow:

\begin{equation}
s \in C^h \Leftrightarrow \phi(p(C^h)) = \max_{l \in \{1,...,K\}} \{\phi(p(C^l))\}
\label{eq.decisionrule}
\end{equation}
The general framework of the proposed \textit{CR} classifier under 
leave-one-out cross-validation (LOOCV) is described below.

\begin{itemize}

\item \textbf{Step 1 - Data Partitioning:}  
Given a dataset of $n$ samples, split the data into $n$ folds, 
where each fold contains exactly one sample.

\item \textbf{Step 2 - Leave-One-Out Iterative Procedure:}  
Repeat the following process for each fold:
\begin{itemize}
    \item Select one fold (one sample) as the testing set.
    \item Use the remaining $(n-1)$ samples as the training set.
\end{itemize}

\begin{itemize}
\item \textit{Phase 1 - Training Phase:}
\begin{itemize}
    \item \textit{Class Separation:}  
    Partition the training set into classes according to the target variable.

    \item \textit{Feature Interval Estimation:}  
    For each class and each feature, compute representative feature intervals using statistical descriptors (e.g., mean and standard deviation in equation\ref{eq.meanstd} or percentile-based boundaries in equation \ref{eq.percentile}), resulting in lower and upper bounds for each feature.

    \item \textit{Prototype Construction:}  
    Construct a class prototype by aggregating the feature intervals for all features within each class.
\end{itemize}

\item \textit{Phase 2 - Testing Phase:}
\begin{itemize}
    \item \textit{Partial Distance Computation:}  
    Compute the partial distance between the test sample and each class prototype for every feature (equation \ref{eq.absdistance}).

    \item \textit{Partial Resemblance Evaluation:}  
    Convert the partial distances into partial resemblance values between the test sample and each class prototype (equation \ref{eq.partialoutranking}).

    \item \textit{Weighted Resemblance Aggregation:}  
    Combine the partial resemblances using predefined feature weights to form a weighted resemblance score for each class (equation \ref{eq.globalOutranking}).

    \item \textit{Global Resemblance Score:}  
    Compute a global resemblance score for each class by aggregating the weighted feature-wise resemblances (equation \ref{eq.netflow}).

    \item \textit{Decision Rule:}  
    Assign the test sample to the class that achieves the highest global resemblance score (equation \ref{eq.decisionrule}).
\end{itemize}
\end{itemize}

\item \textbf{Step 3 - Performance Evaluation:}  
Record the prediction outcome and compute the selected performance metrics for the test sample.

\item \textbf{Step 4 - Iteration Over All Samples:}  
Repeat Steps 2 and 3 until each sample has been used once as the test sample.

\item \textbf{Step 5 - Final Performance Estimation:}  
Compute the average of the performance metrics across all $n$ iterations to obtain the overall classification performance.

\end{itemize}

The source code of the CR classifier is available on GitHub at [\url{https://github.com/nbelacel/Closest-Resemblance}].

The classification decision rule by CR yields an interpretable decision process, balancing feature-level resemblance and global class similarity. The use of interval learning and outranking relations provides robustness to noise, measurement uncertainty, and imbalanced distributions conditions \citep{belacel_alg2025, belacelREMOTESensingData2020k} frequently encountered in biomedical datasets such as ADHD metabolomics analysis.

\subsection{Proposed feature selection approach}
Feature selection is a central challenge in machine learning applied to metabolomics, where high-dimensional biochemical data often contain substantial redundancy and noise. In the context of ADHD biomarker discovery, selecting the most informative metabolites is essential for improving diagnostic accuracy, preventing overfitting, and enhancing the biological interpretability of predictive models \citep{Cao2020MetabML, Alonso2015MetaboMLReview}. Prior work in psychiatric metabolomics demonstrates that robust feature selection can reveal biologically meaningful metabolic disruptions—such as alterations in amino acid metabolism, catecholamine turnover, and energy pathways while reducing the likelihood of identifying spurious markers driven by cohort-specific variation \citep{tian2022urinary, Liu2019ADHDMetab}. As metabolomic datasets for ADHD typically involve dozens to hundreds of quantified metabolites, effective feature-selection strategies remain critical for developing reproducible, clinically scalable diagnostic models \citep{Castro2023PsychMetabReview}.
Feature selection techniques are commonly grouped into three categories: filter, wrapper, and embedded approaches \cite{Jovic2015}. Filter methods evaluate the relevance of features independently from the classifier, using statistical or correlation-based criteria. Wrapper methods, in contrast, rely on the performance of a specific classifier to assess feature subsets, requiring the model to be trained iteratively. Embedded methods integrate feature selection directly into the model training process, as is the case for classifiers that adjust feature weights or intervals during learning.

In this study, we implemented a hybrid feature selection approach specifically adapted for the CR classifier. Starting from the full set of features, we employed a greedy backward elimination strategy combined with Monte Carlo–based weight optimization to identify the most discriminative subset of features (\cite{Belacel2019}). The procedure begins by applying the CR classifier with uniform feature weights to establish a baseline classification accuracy ($f$). Then, a Monte Carlo simulation generates random weight vectors for all features within a normalized $[0,1]$ range that sums to one. The CR classifier is repeatedly applied using these weight vectors, and any configuration achieving an improved classification accuracy ($f' \geq f$) is retained. Features with weights below a defined threshold ($t_0$) are iteratively removed, yielding a reduced feature subset ($SF'$). The process continues until no further improvement is observed or a predefined computational limit is reached. The algorithm \ref{alg:CRFeatureSelection} summarizes the different steps of our feature selection approach.
\begin{algorithm}[!ht]
\caption{Feature Selection Procedure for the Closest Resemblance (CR) Classifier}
\label{alg:CRFeatureSelection}
\begin{algorithmic}[1]
\Require Feature set $\mathcal{F} = \{f_1, f_2, \dots, f_m\}$; threshold $t_0$; maximum iterations $N_{max}$
\Ensure Reduced feature subset $SF^*$ and weight vector $W^*$
\State Initialize feature subset $SF \gets \mathcal{F}$ and equal weights $w_i = 1/m$
\State Apply CR classifier with $SF$ to compute baseline accuracy $A_0$
\For{$iter = 1$ to $N_{max}$}
    \State Generate random weights $w_i \in [0,1]$ such that $\sum w_i = 1$ \Comment{Monte Carlo sampling}
    \State Apply CR classifier using weights $W = (w_1, \dots, w_{|SF|})$
    \State Compute new accuracy $A'$
    \If{$A' \geq A_0$}
        \State $A_0 \gets A'$, $W^* \gets W$
    \EndIf
    \State Remove features $f_i$ with $w_i \leq t_0$ to obtain reduced subset $SF'$
    \State $SF \gets SF'$
    \If{No improvement in $A_0$ or $|SF|$ unchanged}
        \State \textbf{break}
    \EndIf
\EndFor
\State \Return $SF^*$ and $W^*$
\end{algorithmic}
\end{algorithm}

The embedded feature-selection mechanism iteratively identifies a pseudo-optimal subset of urinary metabolites, ranked according to their final contribution weights, which reflect their relative importance in distinguishing ADHD from control profiles. These weights highlight the metabolites exerting the strongest influence on class separability often those involved in neurotransmitter synthesis, amino acid turnover, or oxidative pathways disrupted in ADHD \citep{tian2022urinary, Liu2019ADHDMetab}. 

\begin{figure}[htbp]
    \centering
    \includegraphics[width=0.8\linewidth]{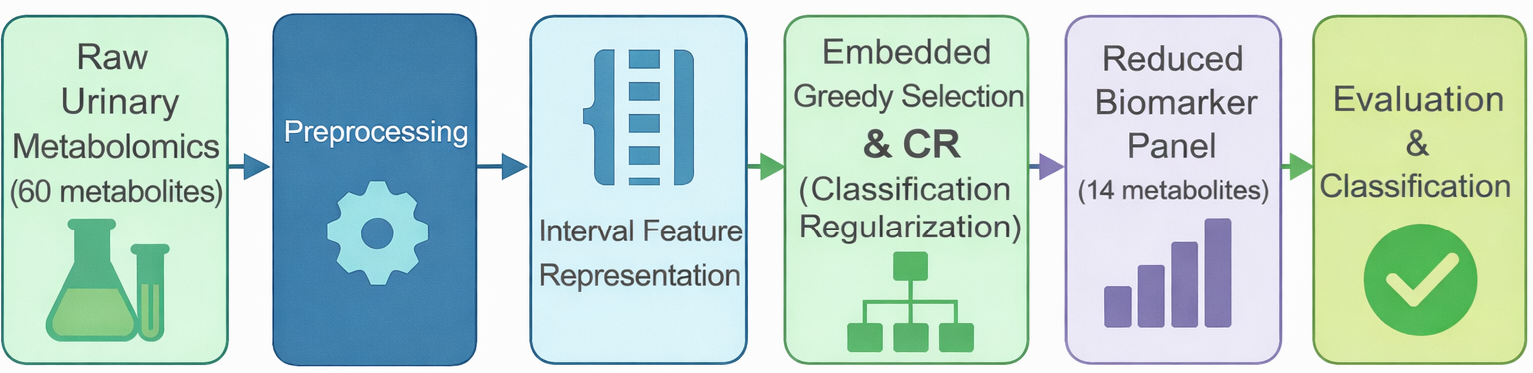}
    \caption{Overview of the proposed Biomarker discovery framework for urinary metabolomic ADHD diagnosis.}
    \label{fig:framework}
\end{figure}
The overall workflow of the proposed feature-selection framework for urinary metabolomic ADHD diagnosis is illustrated in Figure \ref{fig:framework}. The analysis begins with the urinary metabolomics dataset consisting of 60 quantified metabolites from ADHD patients and healthy controls. Following standard preprocessing steps, each metabolite is transformed into an interval-based feature representation reflecting intra-class variability. This interval representation serves as input to the Closest Resemblance (CR) classifier, which evaluates resemblance between samples and class prototypes. An embedded greedy search procedure iteratively selects the subset of metabolites that maximizes CR classification performance while minimizing redundancy. The resulting reduced biomarker panel consists of 14 highly discriminative metabolites biologically linked to neurotransmitter turnover and amino acid metabolism. Finally, the CR classifier and comparison models (RF, kNN) are evaluated on both the full and reduced feature sets to assess predictive accuracy, computational efficiency, and clinical interpretability. This pipeline integrates interval learning and embedded feature selection to identify a concise and biologically meaningful metabolite signature for objective ADHD diagnosis.
Because the search strategy is heuristic rather than exhaustive, the resulting subset represents a near-optimal configuration tailored to the CR classifier. This approach offers an efficient balance between computational tractability and interpretability, a crucial advantage in metabolomics where variables are often highly correlated and biological pathways interact nonlinearly \citep{Cao2020MetabML, Alonso2015MetaboMLReview}. As such, the method is well suited for biomarker discovery pipelines aiming to identify reproducible and clinically meaningful metabolic signatures for ADHD.
\section{Results and Discussion}
\label{sec:results}

\subsection{Classification performance with all metabolites}
\label{subsec:all_metabolites}

We first evaluated the performance of the proposed CR classifier using the full set of 60 urinary metabolites (46 controls, 52 ADHD). Leave-one-out cross-validation (LOOCV) was used to maximize statistical reliability given the cohort size.

When class prototypes were computed using the interquartile range (30th and 80th percentiles), the CR classifier achieved an accuracy of \(\mathbf{95.9\%}\) with AUC \(=0.96\), precision \(=0.96\), and F1-score \(=0.96\) (Table~\ref{tab:perf-full}). Using standard deviation intervals for prototype construction reduced performance slightly (accuracy \(=93.9\%\), AUC \(=0.94\)). Comparative baselines were evaluated under identical validation conditions: Random Forest (500 trees) provided the strongest baseline (accuracy \(=91.8\%\), AUC \(=0.91\)), while KNN and Nearest Centroid performed less well (accuracy range 69–83\%). The CR classifier maintained a very low computational cost (average training time \(<0.01\) s), reflecting its suitability for lightweight deployment.

\begin{table}[!ht]
\centering
\caption{Summary of classifier performance using the full 60-metabolite feature set (LOOCV).}
\label{tab:perf-full}
\begin{tabular}{lcccc}
\toprule
Classifier & Accuracy & AUC & F1-score & Avg. training time (s) \\
\midrule
CR (30–80\%) & \textbf{0.959} & \textbf{0.96} & 0.96 & 0.005 \\
CR (Std. dev.) & 0.939 & 0.94 & 0.94 & 0.007 \\
Random Forest (500 trees) & 0.918 & 0.91 & 0.92 & 61.57 \\
KNN (k=3) & 0.826 & 0.83 & 0.83 & 0.023 \\
Nearest Centroid & 0.694 & 0.71 & 0.68 & 0.033 \\
\bottomrule
\end{tabular}
\end{table}

\subsection{Performance after embedded feature selection}
\label{subsec:feature_selection}

To improve interpretability and reduce dimensionality, an embedded greedy feature-selection procedure within the CR framework was applied. The algorithm selected 14 metabolites as the most discriminant features. Using only these 14 metabolites, the CR classifier (percentile prototypes) achieved accuracy \(=\mathbf{97.9\%}\) and AUC \(=0.978\), while the standard deviation variant achieved accuracy \(=95.9\%\) (Table~\ref{tab:perf-reduced}).

Reducing the feature set from 60 to 14 improved performance for the CR classifier, indicating that the selected metabolites capture the most diagnostic information. Applying the same reduced feature set to other classifiers produced more modest results (e.g., RF accuracy \(=0.908\); KNN\(_{k=3}\) accuracy \(=0.826\)), underscoring that the CR approach is particularly well-matched to metabolomic data where feature correlation and redundancy can degrade traditional methods.

\begin{table}[!ht]
\centering
\caption{Classifier performance using the 14 selected metabolites (LOOCV).}
\label{tab:perf-reduced}
\begin{tabular}{lcccc}
\toprule
Classifier & Accuracy & AUC & F1-score & Avg. training time (s) \\
\midrule
CR (30–80\%) & \textbf{0.979} & \textbf{0.978} & 0.98 & 0.006 \\
CR (Std. dev.) & 0.959 & 0.96 & 0.96 & 0.003 \\
Random Forest (500 trees) & 0.908 & 0.91 & 0.91 & 58.52 \\
KNN (k=3) & 0.826 & 0.82 & 0.83 & 0.040 \\
Nearest Centroid & 0.602 & 0.62 & 0.59 & 0.022 \\
\bottomrule
\end{tabular}
\end{table}

\begin{figure}[!ht]
    \centering
    \includegraphics[width=0.95\linewidth]{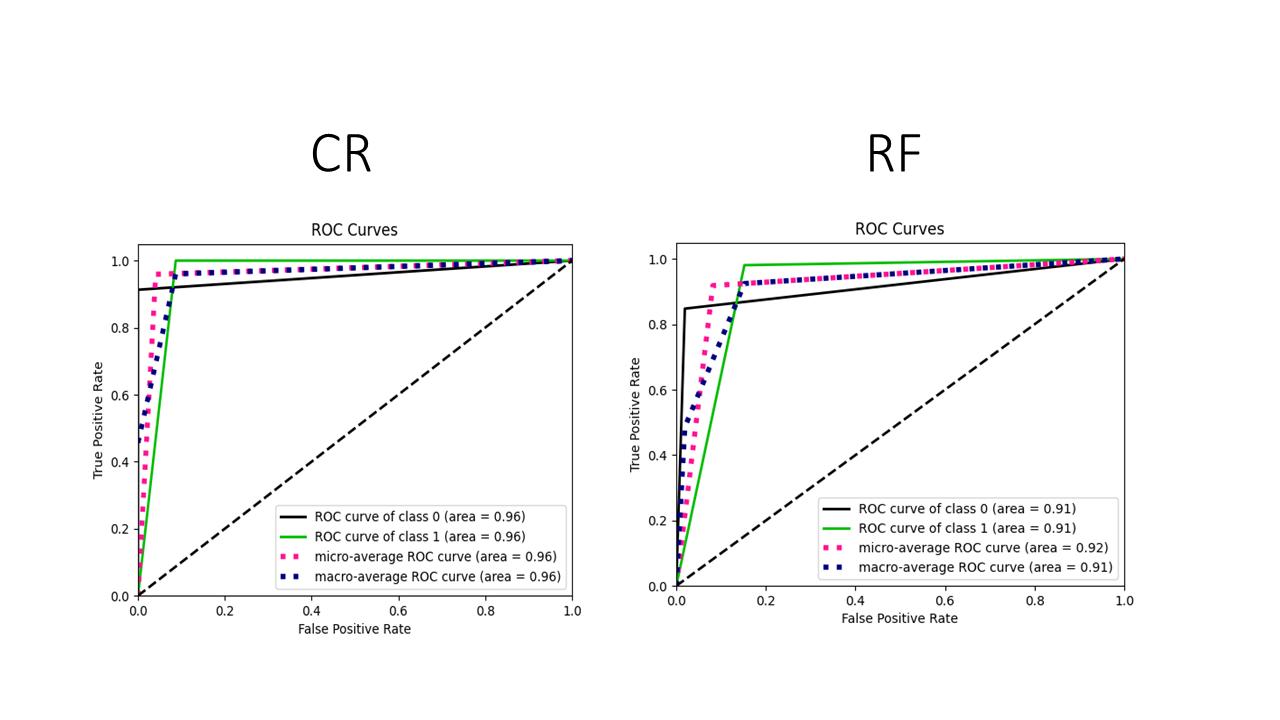}
    \caption{ROC curves comparing the Closest Resemblance (CR) and Random Forest (RF) classifiers using the full set of 60 urinary metabolites. The CR model demonstrates superior discriminative performance (AUC = 0.96) relative to RF (AUC = 0.91).}
    \label{fig:roc-full}
\end{figure}

\begin{figure}[!ht]
    \centering
    \includegraphics[width=0.95\linewidth]{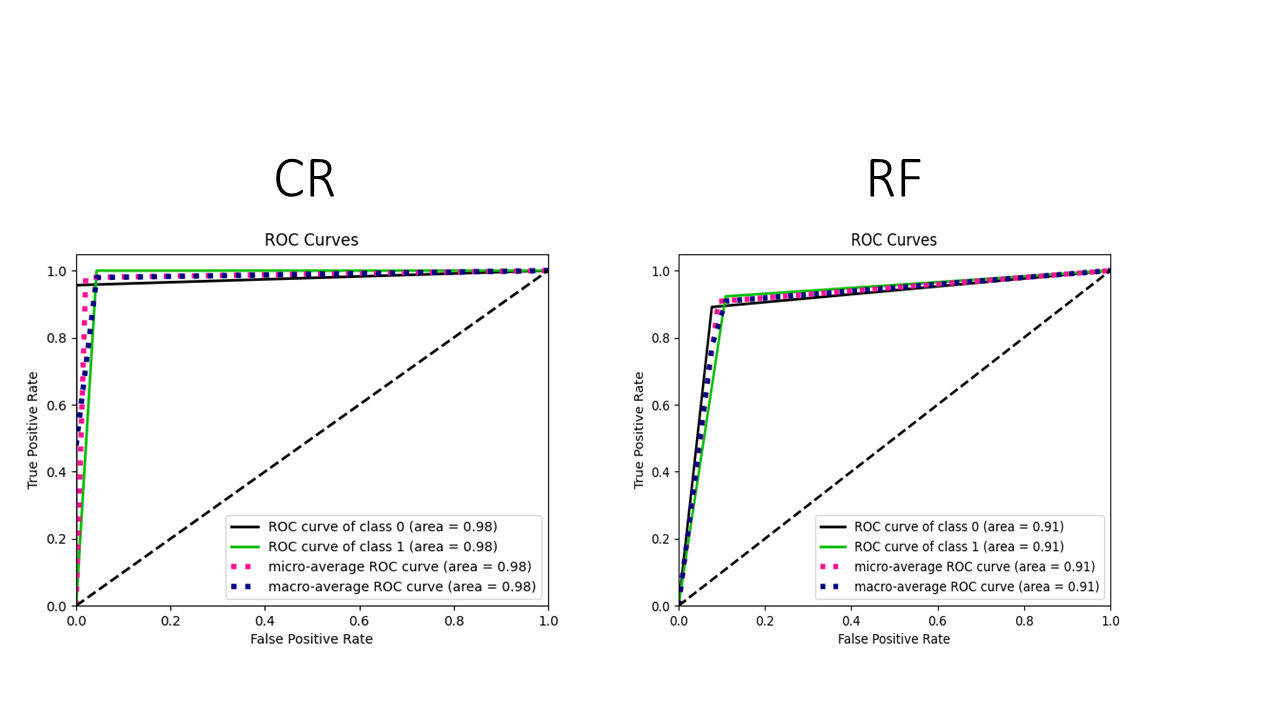}
    \caption{ROC curves for the Closest Resemblance (CR) and Random Forest (RF) classifiers using the reduced metabolite subset obtained from the embedded feature selection procedure. Both models retained high discriminative performance, with CR achieving AUC = 0.98 and RF AUC = 0.91.}
    \label{fig:roc_reduced}
\end{figure}

\subsection{Receiver Operating Characteristic (ROC) analysis}
\label{subsec:roc}

To further evaluate the discriminative capability of the proposed CR classifier, Receiver Operating Characteristic (ROC) curves were generated and compared with those obtained from the Random Forest (RF) model (Figure~\ref{fig:roc-full} and Figure~\ref{fig:roc_reduced}). As shown in Figure~\ref{fig:roc-full}, the CR classifier using the full set of 60 urinary metabolites achieved an Area Under the Curve (AUC) of 0.96, outperforming the RF model (AUC = 0.91). This indicates superior sensitivity–specificity trade-offs and robustness in distinguishing ADHD from control subjects.

When the models were applied to the reduced subset of metabolites identified by the embedded feature selection procedure, both classifiers retained high discriminative power, with the CR classifier maintaining an AUC of 0.98 compared to 0.91 for RF (Figure~\ref{fig:roc_reduced}). 
The results clearly show that the proposed feature selection strategy enhances classification accuracy and efficiency across all tested models. When using the fourteen selected features, the CR (lower percentile=0.3; upper percentile = 0.8) classifier achieved the highest overall performance, with an AUC of 0.98, and an F1-score of 0.98. In contrast, using all 60 features yielded lower values (AUC = 0.96, and F1 = 0.96). Similarly, the CR (t=0.3) variant improved from an AUC of 0.94 to 0.96 after feature reduction, confirming the stability and robustness of the Closest Resemblance framework.
The minimal decline in AUC confirms that the selected metabolite subset preserves most of the relevant diagnostic information while reducing model complexity.

From a translational perspective, these findings highlight that the CR classifier not only achieves competitive accuracy but also maintains interpretability and computational efficiency, making it a suitable candidate for integration into metabolomic-based diagnostic workflows or edge-AI clinical systems.

\subsection{Biological relevance of selected metabolites}
\label{subsec:biological}

The 14 metabolites selected by the CR greedy algorithm include several compounds with high biological plausibility in the context of ADHD. Notably, metabolites connected to catecholamine metabolism (e.g., Dopamine 4-sulfate, 4-Methoxytyramine) and amino-acid/urea cycle intermediates (e.g., Citrulline) were retained. These findings align with prior metabolomic analyses reporting alterations in neurotransmitter metabolism and amino acid pathways in ADHD \citep{tian2022urinary, cortese2008, macDonald2024}.  

Additionally, the selection included metabolites potentially linked to polyamine and tryptophan metabolism (e.g., N-Acetylisoputreanine, Indolylacryloylglycine). While some of these molecules are less widely reported in ADHD literature, they may indicate understudied metabolic axes relevant to ADHD pathophysiology and merit targeted biochemical validation.

\subsection{Clinical Utility and Potential Impact}
\label{subsec:impact}
The findings of this study have meaningful implications for clinical practice, particularly in efforts to develop objective, scalable, and biologically grounded diagnostic tools for ADHD. From a translational perspective, our results suggest that urinary metabolomic profiles can support non-invasive ADHD screening. The 14-metabolite panel identified by the CR classifier can be quantified using targeted LC–MS/MS assays that are increasingly available in clinical laboratories, enabling cost-effective and non-invasive biochemical screening \citep{KaddurahDaouk2009, Emwas2023}. The CR classifier’s interpretability (prototype-based decision rule) and low computational footprint make it suitable for integration into edge-AI or embedded diagnostic platforms, where transparent and low-latency algorithms are preferred \citep{Holzinger2019, Zhang2023}.
While additional external validation is required, these results suggest that interpretable machine learning applied to urinary metabolomics may meaningfully complement current behavioral assessments and enhance precision-medicine approaches in ADHD diagnosis.

\begin{figure}[!ht]
    \centering
    \includegraphics[width=0.95\linewidth]{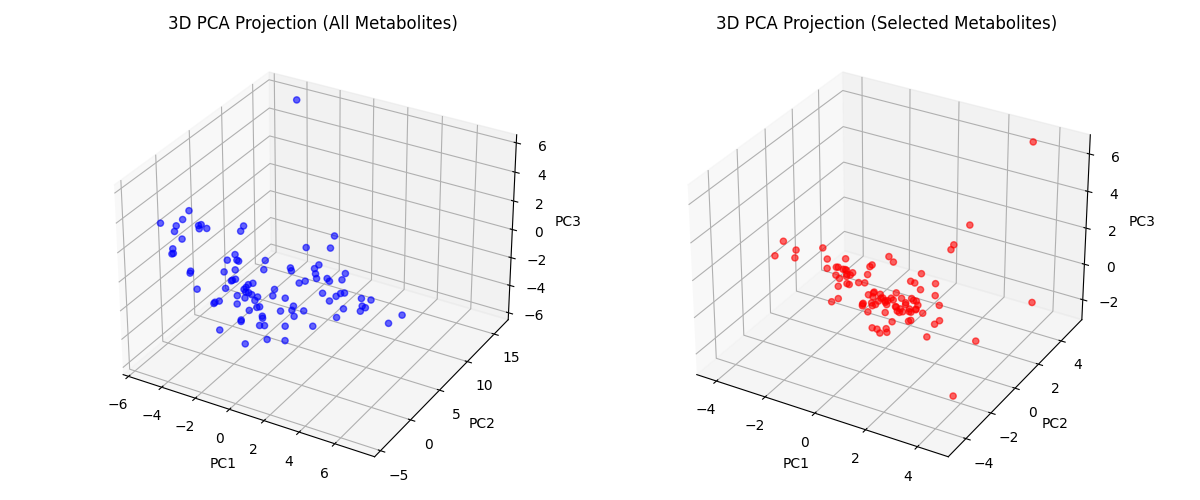}
    \caption{3D PCA projections of the datasets using the full set of 60 metabolites (left) and the reduced set of 14 metabolites obtained through feature interval learning (right). The reduced feature space exhibits more compact and well-separated groupings, indicating that the proposed interval-based feature learning approach effectively removes redundant metabolites while enhancing the discriminative structure of the data.}
    \label{fig:pca}
\end{figure}

\subsection{Limitations and future directions}
\label{subsec:limitations}

Key limitations include the moderate sample size and reliance on a single cohort; independent external validation is required to determine generalizability. Future work will apply the CR framework to external cohorts (e.g., the validation batch reported by \citep{tian2022urinary}) and explore the longitudinal stability of the identified markers for ADHD diagnosis. Finally, targeted quantitative assays should be developed to validate absolute concentration changes for the candidate metabolites.

\section{Conclusion}
Accurate and reliable ADHD assessment requires machine learning models that are not only powerful but also well-adapted to the nature of metabolomic data. The objective of this study was to adapt and apply a machine learning methodology for the selection of a subset of key urinary metabolites that most effectively distinguish ADHD patients from healthy controls.
The proposed approach integrates a feature selection mechanism within the CR classifier framework. By embedding a greedy-based feature selection algorithm into the CR model, the method efficiently identifies the most relevant metabolomic markers while minimizing redundancy. Experimental results demonstrated that this embedded feature selection significantly enhances classification performance for ADHD diagnosis. The CR classifier, when trained on the selected metabolites, consistently outperformed conventional models such as $k$-NN, Random Forest, and Nearest Centroid in terms of predictive accuracy, robustness, and computational efficiency.
Beyond accuracy, the study highlights the advantages of using classifiers based on interval feature learning and outranking principles. These characteristics allow the CR model to better handle data imperfections such as noise, variability, and class imbalance commonly encountered in clinical metabolomic datasets. The simplicity and interpretability of the proposed greedy search algorithm make it particularly well-suited for practical applications where computational resources and transparency are crucial.
The CR with embedded feature selection capability not only improves classification accuracy but also enhances biological interpretability, paving the way for practical clinical translation in precision psychiatry.
Future research will focus on enhancing the feature selection process by integrating metaheuristic optimization strategies, such as variable neighborhood search or particle swarm optimization, to further refine the selection of optimal subsets of metabolite markers. In addition, we plan to validate the identified urinary markers on independent ADHD cohorts to assess their generalizability, translational relevance, and robustness across populations.





\bibliographystyle{elsarticle-num} 
\bibliography{references}





\end{document}